\documentclass[conference]{IEEEtran}
\IEEEoverridecommandlockouts
\usepackage{cite}
\usepackage{amsmath,amssymb,amsfonts}
\usepackage{graphicx}
\usepackage{textcomp}
\usepackage{xcolor}
\def\BibTeX{{\rm B\kern-.05em{\sc i\kern-.025em b}\kern-.08em
    T\kern-.1667em\lower.7ex\hbox{E}\kern-.125emX}}

\usepackage{subcaption}
\usepackage{amsmath}
\usepackage{algorithm}
\usepackage{algpseudocode}
\usepackage{hyperref}
\usepackage{tikz}
\usepackage{float}
 
\usepackage{graphicx}
\usepackage{xspace}

\usepackage{multirow}
\usepackage{enumitem}
\begin{document}
\title{
Federated Split Learning for Human Activity
Recognition with Differential Privacy 
}

\author{
        \IEEEauthorblockN{
        Josue Ndeko,  Shaba Shaon, Aubrey Beal,   Avimanyu Sahoo,  Dinh C. Nguyen
	}

\IEEEauthorblockA{
	Department of Electrical and Computer Engineering, University of Alabama in Huntsville, USA \\
    Emails: josuendeko12@gmail.com, ss0670@uah.edu, aubrey.beal@uah.edu, as0472@uah.edu, dinh.nguyen@uah.edu
	}\vspace{-0pt}}
	\markboth{}%
	{}
\maketitle


\newpage \pagestyle{plain}




\begin{abstract}

This paper proposes a novel intelligent human activity recognition (HAR) framework based on a new design of Federated Split Learning (FSL)  with Differential Privacy (DP) over edge networks. Our FSL-DP framework leverages both accelerometer and gyroscope data, achieving significant improvements in HAR accuracy. The evaluation includes a detailed comparison between traditional Federated Learning (FL) and our FSL framework, showing that the FSL framework outperforms FL models in both accuracy and loss metrics. Additionally, we examine the privacy-performance trade-off under different data settings in the DP mechanism, highlighting the balance between privacy guarantees and model accuracy. The results also indicate that our FSL framework achieves faster communication times per training round compared to traditional FL, further emphasizing its efficiency and effectiveness. This work provides valuable insight and a novel framework which was tested on a real-life dataset.

\end{abstract}
\thispagestyle{empty}
\maketitle


\section{Introduction}



\textcolor{black}{In recent years, the widespread adoption of mobile devices has led to a growing interest in human activity recognition (HAR) using wearable sensors \cite{wang2016comparative}, \cite{lara2012survey}. This area has emerged as a novel research focus within artificial intelligence and pattern recognition \cite{xu2019innohar}. This field intersects artificial intelligence and pattern recognition, with applications ranging from sports activity detection and smart homes to health support and beyond. By leveraging sensor data from devices like smartphones and wearables, HAR enables real-time monitoring and analysis of human activities. This capability is pivotal in enhancing personalized healthcare, improving athletic performance analysis, and developing intelligent environments that adapt to users' needs seamlessly. Modern HAR systems, particularly those powered by deep learning techniques, offer several advantages over traditional methods \cite{zhou2020deep}. They can automatically learn relevant features from raw sensor data, eliminating the need for manual feature engineering and thus improving accuracy and robustness. Moreover, deep learning models can handle complex, nonlinear relationships in the data, leading to better generalization across different users and activity types. Furthermore, HAR contributes to the advancement of human-computer interaction by enabling natural interfaces that respond to users' physical actions and gestures. This technology has the potential to revolutionize how users interact with devices and how devices understand and respond to human behavior in real-world settings. However, the widespread adoption of HAR is constrained by challenges such as privacy concerns associated with personal data collected from users' devices. Addressing these challenges requires innovative approaches in data anonymization, secure data storage, and compliance with regulations like General Data Protection Regulation (GDPR).}


Federated Learning (FL) concepts have risen as potential solutions to privacy concerns. FL is a collaborative framework that takes advantage of user devices' enhanced computational power. In FL multiple users collaboratively train a machine learning model without sharing their personal data with the server, or other users. A global model is downloaded from the server by all user devices, then the users individually train the model using their local data, then finally the server averages the parameters of all users to form a new global model. This process is then repeated until the desired performance has been reached and shared with the client devices. Some models require many parameters which signify a potential limitation of FL.

New concepts involving Split Learning (SL) have been developed as potential solutions to the limitations of FL \cite{turina2021federated},\cite{ha2021spatio}. In SL, a model is initiated and then split into two models, the client-side model and the server-side model. Clients download their side of the model, which involves the input layer and preceding layers until the pre-defined cut layer. The Server-side model contains the rest of the model starting from the cut layer, to the output layer. Training begins at the client-side model with the client's raw data until the cut layer is reached, the intermediate activations at this layer are then sent to the server to continue training. The server then trains its model up to the output layer, computes the loss, and starts back-propagation which will be completed on the client's side.

Federated Split Learning (FSL) has been developed to take advantage of both FL's collaborative framework and SL's splitting structure. The training process in FSL is similar to that in SL until the completion of the backward pass. The weights of the client models are then aggregated to produce a new global client model, and the server-side weights are updated based on the computed gradients. This method efficiently trains a deep neural network model while preserving user data privacy.

\subsection{Related Works}
\textcolor{black}{Several state-of-the-art techniques have introduced feature extraction and selection methods for HAR using traditional machine learning classifiers. With the emergence and advancement of high computational resources, deep learning techniques have become widely used in various HAR systems. These techniques efficiently retrieve features and perform classification, significantly enhancing the performance of HAR systems. Recently, FL has been studied to further improve the performance of HAR. Specifically, the authors in \cite{1} evaluated FL for training a HAR classifier and compared it to centralized learning using two models—a deep neural network and softmax regression—trained on synthetic and real-world datasets. In \cite{2}, an FL system was proposed for HAR, with a perceptive extraction network for feature extraction to improve recognition accuracy. The study in \cite{3} proposed a prototype-guided FL framework for HAR that addresses data issues in real-world environments by efficiently decoupling representation and classifier in heterogeneous FL settings. The work in \cite{4} proposed an FL system for HAR that dynamically learns personalized models by capturing user relationships and iteratively merging model layers based on user similarity. As reported in \cite{5}, the authors designed a 2-dimensional FL framework to address data insufficiency and security in cyber-physical-social systems.}

\textcolor{black}{Recently, SL and SFL has been recently studied in various IoT domains. The authors in \cite{nguyen2021federated} comprehensively surveyed the emerging applications of FL in IoT networks, exploring FL's potential across various IoT services: data sharing, offloading, caching, attack detection, localization, mobile crowdsensing, and privacy as well as security enhancements. In \cite{lin2024efficient}, the authors proposed an efficient SL framework for resource-constrained edge computing systems, combining the benefits of FL and SL. The authors in \cite{ayad2021improving} introduced a modified SL system with an autoencoder and adaptive threshold mechanism, reducing communication and computation overhead in an IoT system with minimal performance loss. The work in \cite{samikwa2022ares} proposed adaptive resource-aware SL for efficient IoT model training, accelerating processes on resource-constrained devices and minimizing straggler effects with device-targeted split points, while adapting to varying network throughput and computing resources. A dynamic FSL framework was developed by the researchers in \cite{samikwa2024dfl} to address data and resource heterogeneity in IoT, enhancing efficiency through resource-aware split computing of deep neural networks and dynamic clustering of training participants. Despite these research efforts, \textit{the application of FSL has not been investigated for HAR in the literature.} Exploring FSL in HAR is crucial as it paves the way for scalable and efficient deployment of personalized activity recognition systems in IoT and wearable technology, catering to individualized user needs while respecting data privacy and security.}









\subsection{Our Key Contributions}
In this paper, we propose a novel collaborative privacy-enhanced HAR framework through the development of an FSL algorithm with differential privacy (DP) \cite{ranaweera2023improving} in edge computing. The contributions of this paper are summarized as follows.

\begin{enumerate}
    \item We present an FSL algorithm for collaborative privacy-enhanced HAR in edge computing. In this framework, edge devices (EDs) are involved to partially perform forward propagation and backpropagation on the client-side models. The remaining workload of the model training will be executed at the edge server through the server-side model, which allows for mitigating the computation burden on resource-constrained EDs. 
    \item Moreover, to enhance privacy protection for activations, a DP mechanism is integrated at EDs to add a mask to hidden the shared information against potential data threats. 
    \item We conduct simulations on real-life HAR datasets to verify the feasibility of our FSL method. Implementation results demonstrate that our approach can significantly enhance the training performance with much lower training latency, compared with traditional training methods. We also validates the impacts of DP in HAR performance under different training settings. 
\end{enumerate}
\subsection{Paper Structure}

The rest of the paper is structured as follows. In Section \ref{Sec:SystemModel}, we present our system model, detailing the architecture and components of our proposed system. Section 
 \ref{Sec:SimulationsAndEvaluations} presents simulations and performance evaluations for the proposed FSL approach under different network settings. Finally, Section  \ref{Sec:Conclusion} concludes the paper. 

\section{System Model} \label{Sec:SystemModel}

\subsection{Overview}

\begin{figure}[ht!]
    \centering
    \footnotesize
    \begin{subfigure}[b]{0.98\linewidth} 
        \centering
        \includegraphics[width=\linewidth]{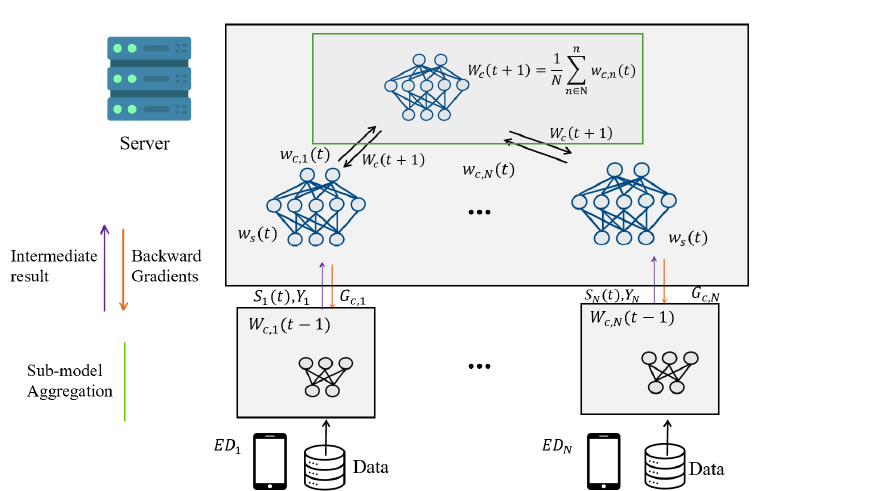}
        \caption{\footnotesize FSL.}
        \label{fig1a}
    \end{subfigure}
    \vspace{1em} 
    \begin{subfigure}[b]{0.98\linewidth} 
        \centering
        \includegraphics[width=\linewidth]{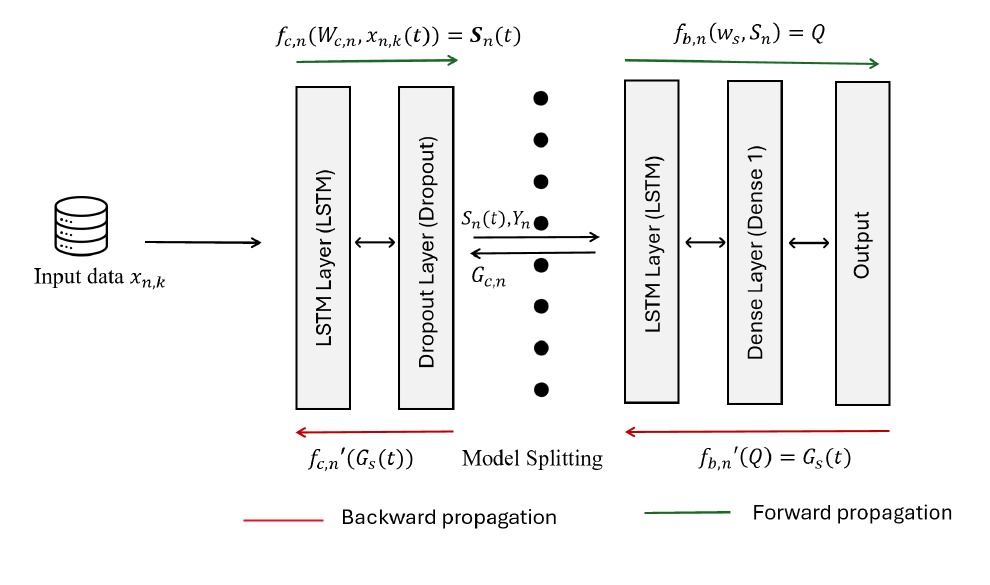}
        \caption{\footnotesize Model Splitting.}
        \label{fig1b}
    \end{subfigure}
    \caption{\footnotesize Proposed FSL framework for HAR over distributed egde networks. Each HAR model is divided into a device-side model and a server-side model. EDs collaborate to train the device-side model and share feature representations to the server that assists to complete the model training by executing the server-side model. }
    \label{fig:overview}
\end{figure}

This paper considers a collaborative HAR system with a set of edge devices (EDs)  denoted by $\mathcal{N}$ and a single edge server, as illustrated in Fig.~\ref{fig:overview}. Each Mobile Device (MD) $n \in \mathcal{N}$ is assumed to be equipped with a wearable sensor attached to the human body and synchronized to emit human motion data. For example, a smartwatch is typically equipped with an accelerometer and a gyroscope which could generate a signal vector at a time. A multidimensional time series containing signal vectors can be used to represent the sensing data over time in HAR.


There are two key components in our collaborative HAR.
\begin{itemize}
    \item EDs: We assume that each client can perform forward propagation and backpropagation on the client-side model using their powerful MD. The local dataset \(D_n\), containing \(D_n\) data samples owned by edge device \(n\), is represented as \(D_n = \{(x_{n,k}, y_{n,k})\}_{k=1}^{D_n}\), where \(x_{n,k} \in \mathbb{R}^{d \times 1}\) denotes the \(k\)-th input data in the local dataset \(D_n\) and \(y_{n,k} \in \mathbb{R}^1\) is the label of \(x_{n,k}\). Consequently, the total dataset \(D\), with \(D = \sum_{n=1}^{N} D_n\) data samples, is denoted by \(D = \bigcup_{n=1}^{N} D_n\). The client-side model is represented as \(W_c \in \mathbb{R}^{u \times 1}\), where \(u\) denotes the dimension of the client-side model’s parameters.
    \item Edge Server: The Edge Server is a computationally powerful system that assesses server-side training. The server-side model is represented as \(W_s \in \mathbb{R}^{r \times 1}\), where \(r\) indicates the dimension of the server-side model’s parameters.
\end{itemize}


\subsection{Proposed FSL Design  for Collaborative HAR}
This section presents the proposed FSL design for a collaborative HAR system. The foundational idea of FSL begins with the server initializing a Machine Learning (ML) model, splitting the model via layer-wise model partitioning, and then sharing the client-side model with all participating MDs. Training is performed next until the model converges. For device $n$, for training round \( t \in T = \{1, 2, \ldots, T\} \), the FSL training process consists of the following stages.

\begin{enumerate}
    \item \textbf{Client-side Model Forward Propagation:} First, forward propagation is performed on the ED for the client-side model. Precisely, a mini-batch \(B_n \subseteq D_n\) containing \(b\) data samples is randomly selected from its local dataset. The input data and labels of the mini-batch are represented by \(X_n(t) \in \mathbb{R}^{b \times d}\) and \(y_n(t) \in \mathbb{R}^{b \times 1}\), respectively, while \(W_{c,n}(t - 1)\) denotes the client-side model of edge device \(n\). After processing the input data from the mini-batch through the client-side model, the cut layer produces the intermediate activations. Intermediate activations from the ED \(n\) are given as the following equation where activation’s dimension per data sample is denoted by \(q\): 
    \begin{equation}
S_n (t) = f \left(X_n (t); W_{c,n} (t - 1)\right) \in \mathbb{R}^{b \times q}
\end{equation}
    \item \textbf{Activations Transmission:} The intermediate activations $S_n (t)$ produced by the client-side models of all participating EDs $\forall n \in \mathcal{N}$ are then shared with the server over a wireless channel. To further enhance privacy for FSL training, we integrate a Gaussian-based local Renyi DP mechanism $\mathcal{G}(0,\alpha_n)$ for EDs. Each ED $n$ enforces DP to its activations by adding a certain amount of noise defined by a pair of parameters $(\epsilon_n, \alpha_n)$, which provides a strong criterion for privacy preservation. Here, $\alpha_n > 0$ is the distinguishable bound of all outputs on neighboring datasets $x,x'$ in a database $X$, and $\alpha_n$ represents the probability of the event that the ratio of the probabilities for two adjacent datasets $x,x'$ cannot be bounded by $e^\epsilon_n$ after adding a privacy-preserving mechanism. With an arbitrarily given $\alpha_n$, a smaller $\epsilon_n$ gives a less distinguishability of neighboring datasets and thus increases the privacy preservation but degrades the training performance, e.g., accuracy. Based on our RDP analysis in previous work \cite{ranaweera2023improving}, we consider the standard deviation of the Gaussian noises, denoted as $\zeta_n$, which is defined as
\begin{equation}
\zeta_n  = \frac{H}{\sqrt{\epsilon_n-z}},
\end{equation}
where $H,z$ are constant. Hence, the activations of each ED $n$ after being enforced by DP can be expressed as 
\begin{equation}
 S_n (t) \leftarrow S_n (t) +  \zeta_n,
\end{equation}
which is then shared with the server for the model forward process.

    \item \textbf{Server-side Model Forward and Back Propagation:} After receiving the DP-enforced activations from the cut layer of ED $n, \forall n \in \mathcal{N}$, the server computes the remaining layers of the model and reaches an output. This marks the conclusion of the forward propagation process. The server then starts the Back Propagation process by calculating the loss using the prediction and the true output. The server continues to compute backpropagation for its layers until the cut layer is reached. The server-side model is expressed as:
\begin{equation}
W_s (t) = \left[ w_s^{\ell_s} (t); w_s^{\ell_s-1} (t); \ldots; w_s^1 (t) \right]
\end{equation}
where \(w_s^k\) is the \(k\)-th layer in the server-side model and \(\ell_s\) signifies the number of server-side model layers. Mathematically, the gradients of the server-side model are expressed as:
\begin{equation}
G_s (t) = 
\begin{bmatrix}
g_s^{\ell_s} (t) \\
g_s^{\ell_s-1} (t) \\
\vdots \\
g_s^1 (t)
\end{bmatrix}
\end{equation}

where \(g_s^k\) is the gradient of the server-side \(k\)-th layer.
    \item \textbf{Client-side Model Back Propagation:} At this stage, each ED uses the received activations to complete the parameter update of its client-side model. The client-side gradients for edge device \(n\) can be calculated as:
\begin{equation}
G_{c,n} (t) =
\begin{bmatrix}
g_{c,n}^{\ell_c} (t) \\
g_{c,n}^{\ell_c-1} (t) \\
\vdots \\
g_{c,n}^1 (t)
\end{bmatrix}
\end{equation}
where \(g_{c,i}^k\) represents the gradients of the client-side \(k\)-th layer of edge device \(n\). The final step for this stage is to update the client-side model. Thus, for ED \(n\), the client-side model is updated using
\begin{equation}
W_{c,n} (t) \leftarrow W_{c,n} (t - 1) - \eta_c G_{c,n} (t)
\end{equation}

    \item \textbf{Aggregation of client model weights:} \textcolor{black}{The final stage of the FSL training process is aggregating the client model weights through federated averaging.
    \begin{equation}
        W_{c} (t+1) = \frac{1}{N} \sum_{n \in N} W_{c,n} (t)
    \end{equation}}
\end{enumerate}


The proposed FSL training framework is outlined in~\textbf{Algorithm~\ref{FSL_procedure}}. Specifically, training is initiated for a predetermined amount of rounds $T$ (line 2). Then for ED $n$, forward propagation is complete on the client model up to the cut layer, and noise is added to the intermediate activations (lines 6 \& 7). The intermediate activations from the client-side model are then processed through the server-side model, producing an output (lines 10 - 12). Once this process is completed for $N$ ED (lines 5 - 13), the loss and gradients of the server-side model are calculated (lines 16 - 18). The client weights are aggregated using Federated Averaging and sent to the clients, along with unaggregated activation gradients (lines 19 - 21). Then finally, the gradients of the client-side model are calculated for all ED.

\begin{algorithm}[ht]
  \caption{The FSL training framework}\label{FSL_procedure}
  \begin{algorithmic}[1]
    \Require
        $b$, ${\eta}_c$, ${\eta}_s$, $\mathcal{N}$, $\cal{D}$, $\phi$, $f_k$ and $B_k$.
    \Ensure
        ${{\bf{W}}^{\bf{*}}}$
    \State Initialization: ${{\bf{W}}}\left( 0 \right)$
    \For{$t=1,2,...,T$ }
    \State
    \State /* Runs on edge devices */
        \ForAll {ED ${n \in \,\mathcal{N}}$ in parallel}
        \State ${{\bf{S}}_n}\left( t \right) \leftarrow f\left( {{{\bf{X}}_n}\left( t \right);{{\bf{W}}_{c,n}}\left( t-1 \right)} \right)$
        \State Add noise: $S_n(t) \leftarrow S_n(t) + \zeta_n$
        \State Send $\left( {{\bf{S}}_n\left( t \right),{\bf{y}}_n\left( t \right)} 
        \right)$ to the server 
        \State /* Runs on server */
        \State ${\bf{S}}\left( t \right) \leftarrow \left[ {{{\bf{S}}_1}\left( t \right);{{\bf{S}}_2}\left( t \right);...;{{\bf{S}}_C}\left( t \right)} \right] $
        \State ${\bf{y}}\left( t \right) \leftarrow \left[ {{{\bf{y}}_1}\left( t \right);{{\bf{y}}_2}\left( t \right);...;{{\bf{y}}_C}\left( t \right)} \right] $
        \State ${\bf{\hat y}}\left( t \right) \leftarrow f\left( {{\bf{S}}\left( t \right);{{\bf{W}}_s}\left( t-1 \right)} \right) $
        \EndFor
    \State
        \State /* Runs on server */
        \State Calculate loss function value $L\left( {{\bf{W}}\left( {t - 1} \right)} \right)$
        \State Calculate gradients of server-side model  ${\bf{G}}_s\left( t \right)$
        \State ${{\bf{W}}_s}\left( t \right) \leftarrow {{\bf{W}}_s}\left( t-1 \right) - {\eta}_s {{\bf{G}}_s}\left( t \right)$
        \State Aggregate client weights:
        \Statex \[
            {\bf{W}}_{c} (t+1) = \frac{1}{N} \sum_{n \in \mathcal{N}} {\bf{W}}_{c,n} (t)
        \]
        \State  Share aggregated weights to all EDs
        \State Send unaggregated activations' gradients to correspo-
        \Statex $\;\;\;\;\;\;$nding ED
    \State
    \State /* Runs on edge devices */
        \ForAll {ED ${n \in \,\mathcal{N}}$ in parallel}
        \State Calculate gradients of client-side model ${{\bf{G}}_{c,n}}\left( t \right)$
        \State ${{\bf{W}}_{c,n}}\left( t \right) \leftarrow {{\bf{W}}_{c,n}}\left( t-1 \right) - {\eta _c}{{\bf{G}}_{c,n}}\left( t \right)$
        \EndFor
    \EndFor
  \end{algorithmic}
\end{algorithm}
\section{Simulations and Evaluation} \label{Sec:SimulationsAndEvaluations}
 \subsection{Simulation Settings}
 
 \textbf{{Dataset Preparation:}}
    For our simulations, we used the \textit{UCI HAR Dataset} \cite{anguita2013public}. The dataset contains 6 regular activities, more specifically 3 dynamic activities (going upstairs, going downstairs, walking) and 3 static activities (standing, sitting, laying). The data was obtained from 30 subjects executing the daily activities listed above. During their performance, the subjects were wearing a Samsung Galaxy S II, using the smartphone's accelerometer and gyroscope, three-axial linear acceleration and three-axial angular velocity at a constant rate of 50 Hz were collected. The collectors then manually labeled the data using the video-recorded. Finalizing their dataset, they randomly divided the data into two sets where 70\% generates the training data, and 30\% the test data. Noise filters were then applied to the sensor data, which were then sampled in fixed-width sliding windows of 2.56 sec and 50\% overlap (128 readings/window)
    
 \textbf{{Training Settings:}}
We carried out experiments on a server equipped with a Core(TM) i7-8700 CPU and 16 GB of RAM. Our algorithm was constructed using PyTorch 2.2.0, and we evaluated the latency per round using Python's time module. To evaluate the algorithm, we ran tests using different $\epsilon$ values, and we also conducted schemes using different data settings. Finally, we compared our FSL model's performance with that of the traditional FL models.

We employed the UCI HAR Dataset for human activity recognition, ensuring a comprehensive evaluation. The dataset includes accelerometer and gyroscope data from 30 subjects, performing six activities: walking, walking upstairs, walking downstairs, sitting, standing, and laying. We used this dataset to evaluate the performance of our FSL framework. We used the LSTM architecture with 100 units and a dropout layer on the client-side network, a dense layer with 100 units, and a softmax output layer on the server-side.

\subsection{Simulation Results}
\subsubsection{Comparison between FSL schemes with and without DP}

\begin{figure}[ht!]
    \centering
    \footnotesize
    \begin{subfigure}[t]{0.49\linewidth} 
        \centering
        \includegraphics[width=\linewidth]{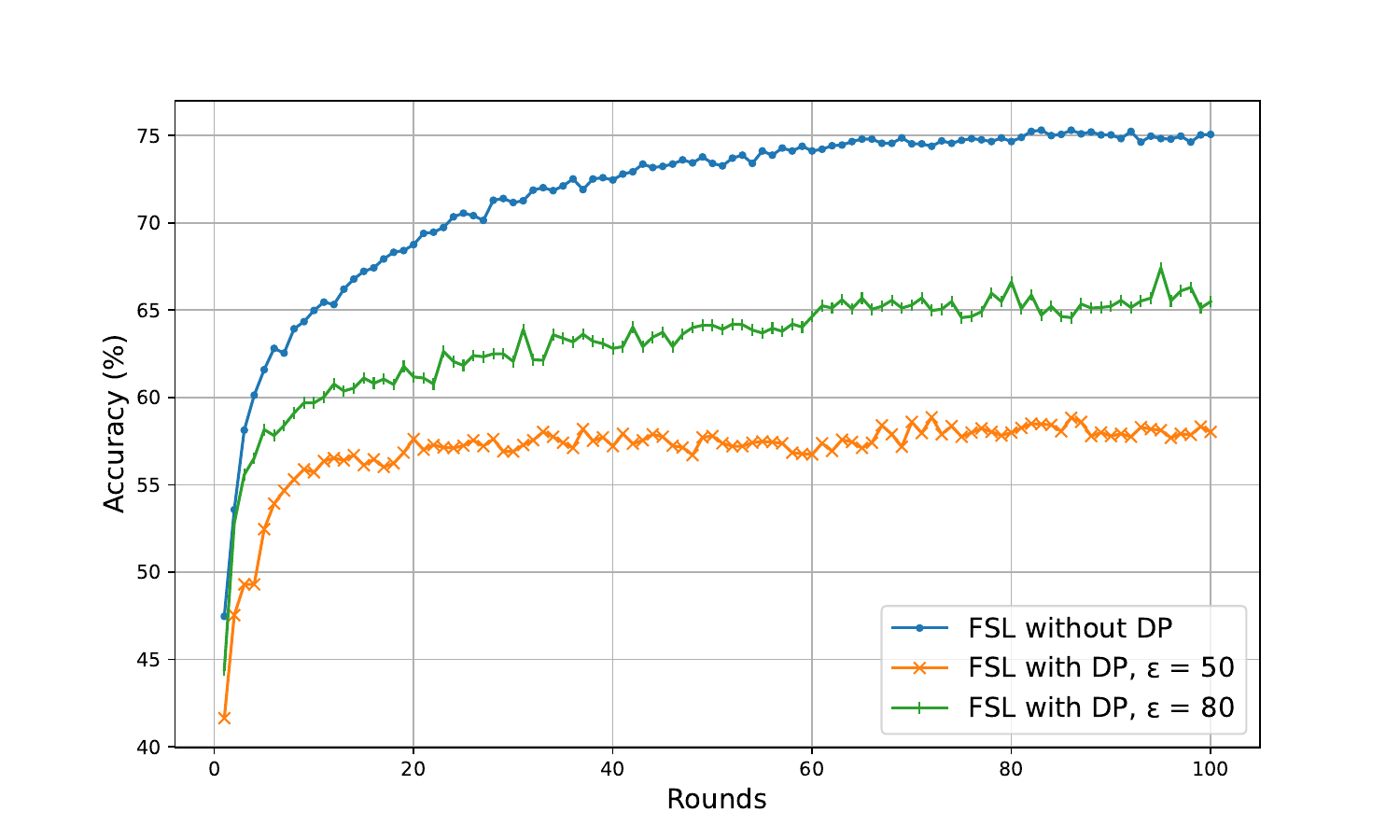}
        \caption{\footnotesize Validation accuracy.}
        \label{fig2a}
    \end{subfigure}
    \hfill 
    \begin{subfigure}[t]{0.49\linewidth} 
        \centering
        \includegraphics[width=\linewidth]{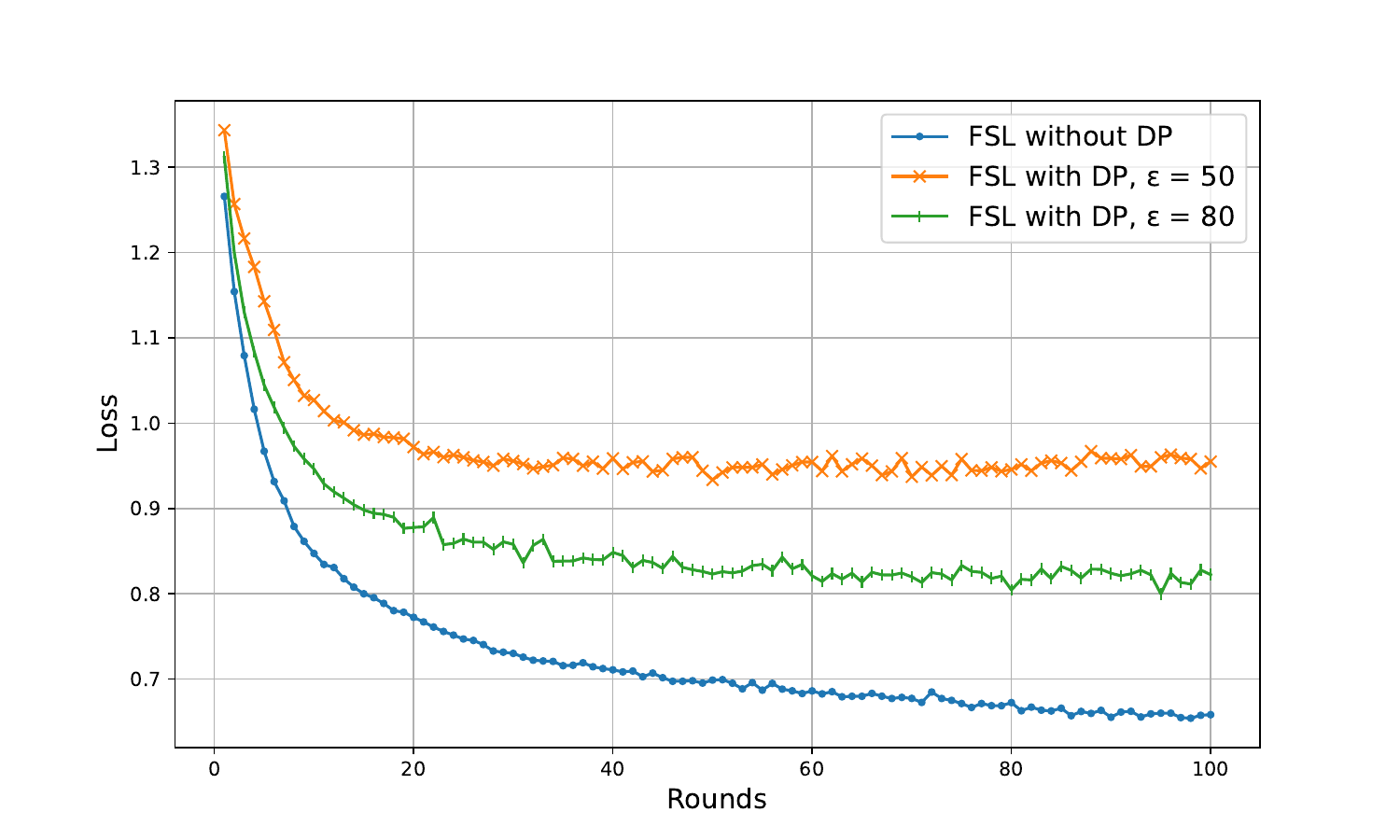}
        \caption{\footnotesize Validation loss.}
        \label{fig2b}
    \end{subfigure}
    \caption{\footnotesize Comparison between FSL schemes with and without DP.}
    \label{fig:three_subs}
\end{figure}

We examine the performance of our FSL model without DP and using various $\epsilon$ values when using DP. Fig.~\ref{fig2a} shows the prominent privacy trade-off as the $\epsilon$ value decreases. Without DP, our model achieves the highest accuracy, indicating that the absence of noise addition allows for more precise model updates. As the $\epsilon$ value decreases, the amount of noise added increases, which degrades our model's accuracy. This is noticeably true when $\epsilon$ = 50 which caused about a 22\% decrease in accuracy compared to the performance without DP, and as $\epsilon$ = 80 cause a lower decrease of about 12\%. In fig.~\ref{fig2b} we observe that DP has a similar effect to our loss value.

\subsubsection{Comparison between FSL schemes under different data settings}

\begin{figure}[ht!]
    \centering
    \footnotesize
    \begin{subfigure}[t]{0.49\linewidth} 
        \centering
        \includegraphics[width=\linewidth]{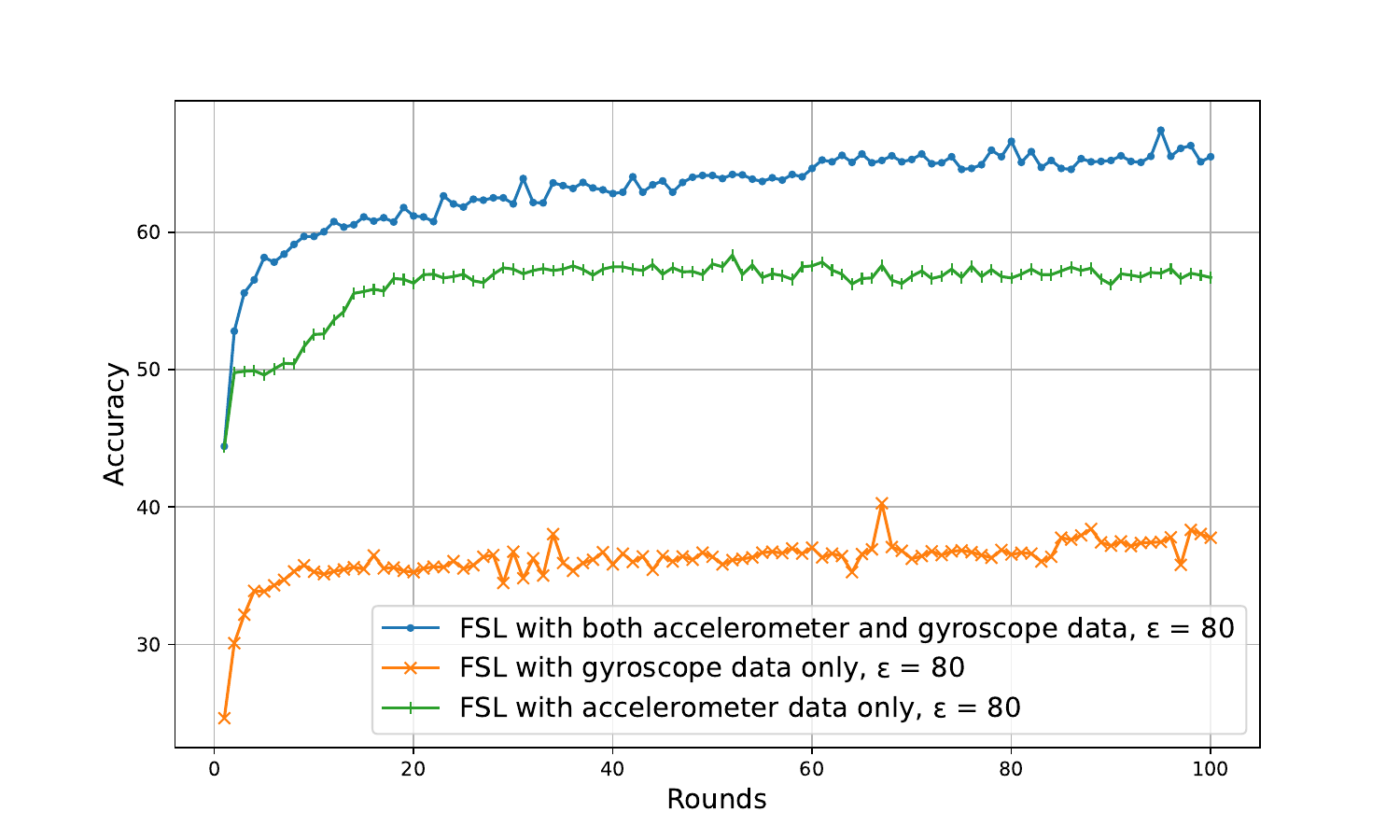}
        \caption{\footnotesize Validation accuracy.}
        \label{fig3a}
    \end{subfigure}
    \hfill 
    \begin{subfigure}[t]{0.49\linewidth} 
        \centering
        \includegraphics[width=\linewidth]{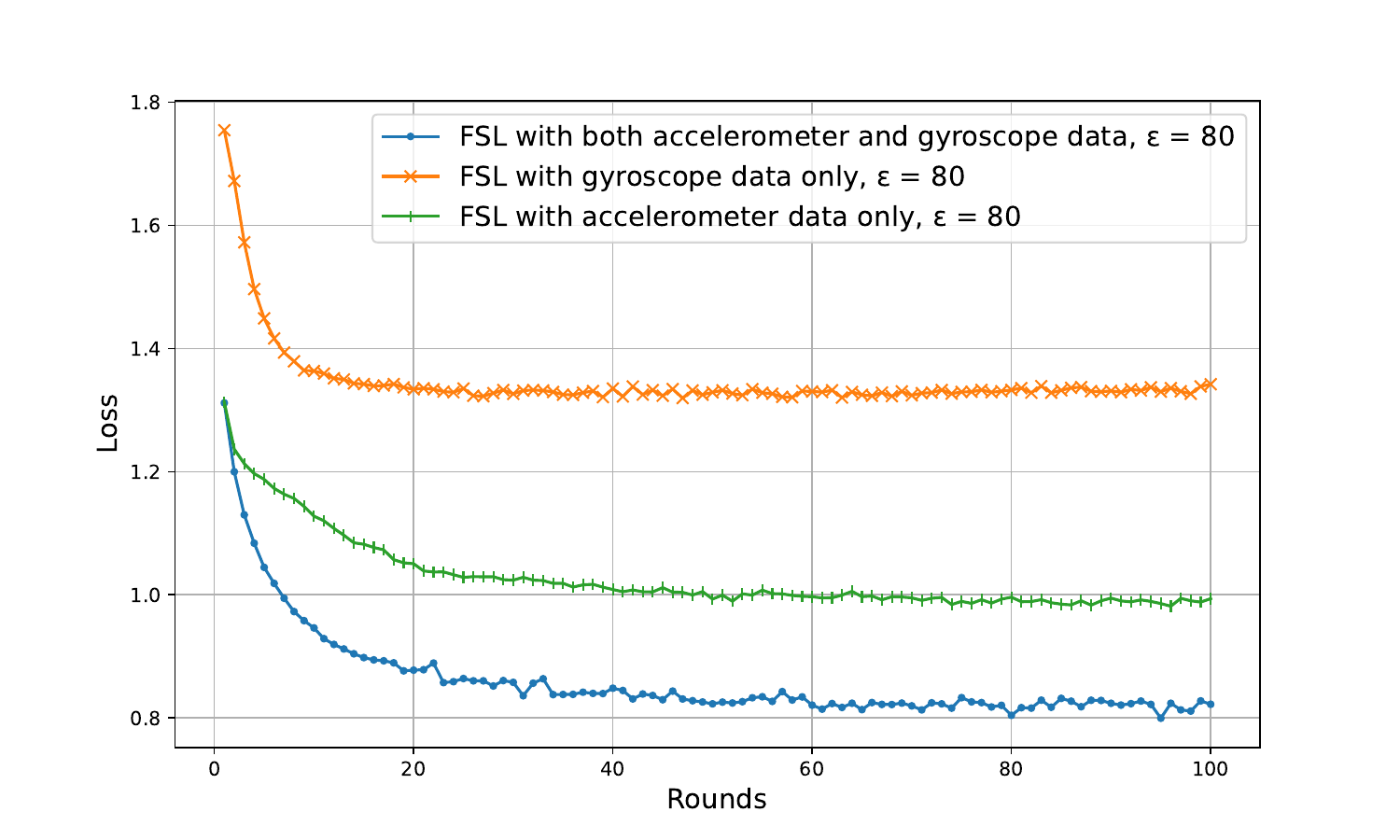}
        \caption{\footnotesize Validation loss.}
        \label{fig3b}
    \end{subfigure}
    \caption{\footnotesize Comparison between FSL schemes under different data settings.}
    \label{fig:three_subs}
\end{figure}

We also examine the accuracy performance of our model under different data settings when $\epsilon$ = 80. Fig.~\ref{fig3a} displays the effect, on accuracy, of using only accelerometer or only gyroscope data as compared to both in our FSL model. Specifically, we notice the following trends: a) when using both gyroscope and accelerometer data, our model achieves the highest accuracy which is a 73.39\% increase from accuracy using gyroscope data only, and a 15.50\% increase from using accelerometer data only. b) As shown in fig.~\ref{fig3b}, our model reached a lower loss value as we used both gyroscope and accelerometer data, indicating more precise predictions and better overall performance.

\subsubsection{Comparison between the proposed FSL and traditional FL}
\begin{figure}[ht!]
    \centering
    \footnotesize
    \begin{subfigure}[t]{0.49\linewidth} 
        \centering
        \includegraphics[width=\linewidth]{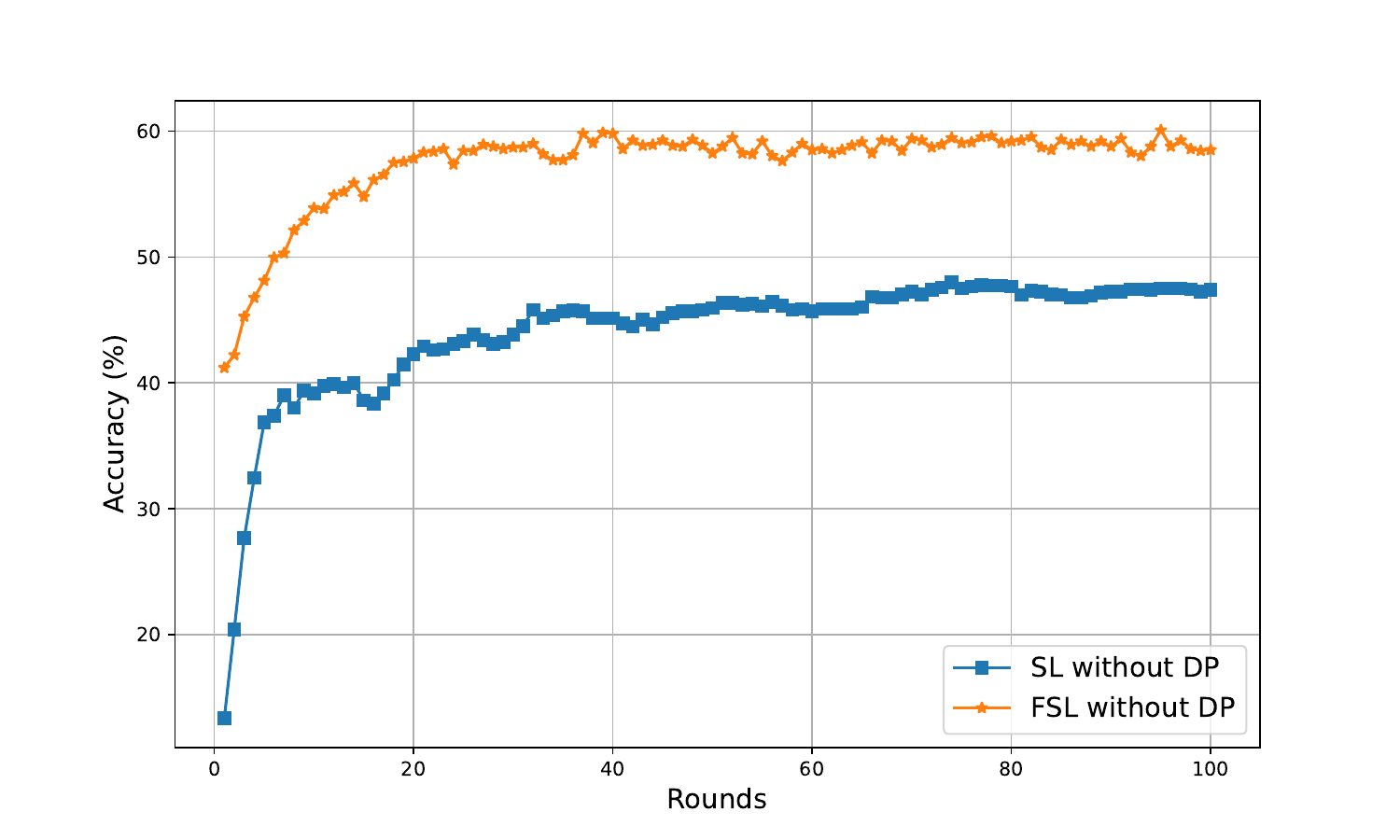}
        \caption{\footnotesize Validation accuracy without DP.}
        \label{fig4a}
    \end{subfigure}
    \hfill 
    \begin{subfigure}[t]{0.49\linewidth} 
        \centering
        \includegraphics[width=\linewidth]{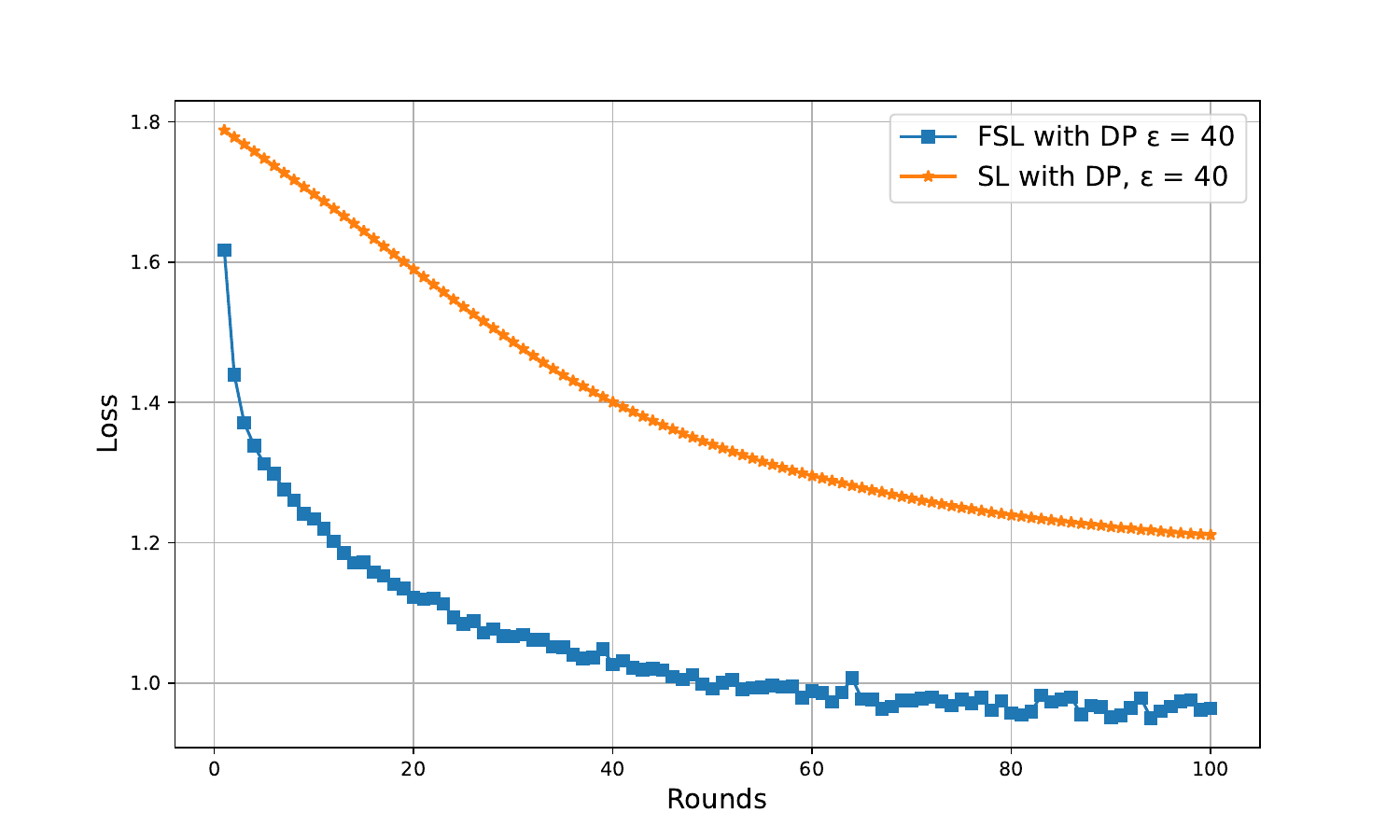}
        \caption{\footnotesize Validation loss without DP.}
        \label{fig4b}
    \end{subfigure}
    \begin{subfigure}[t]{0.49\linewidth} 
        \centering
        \includegraphics[width=\linewidth]{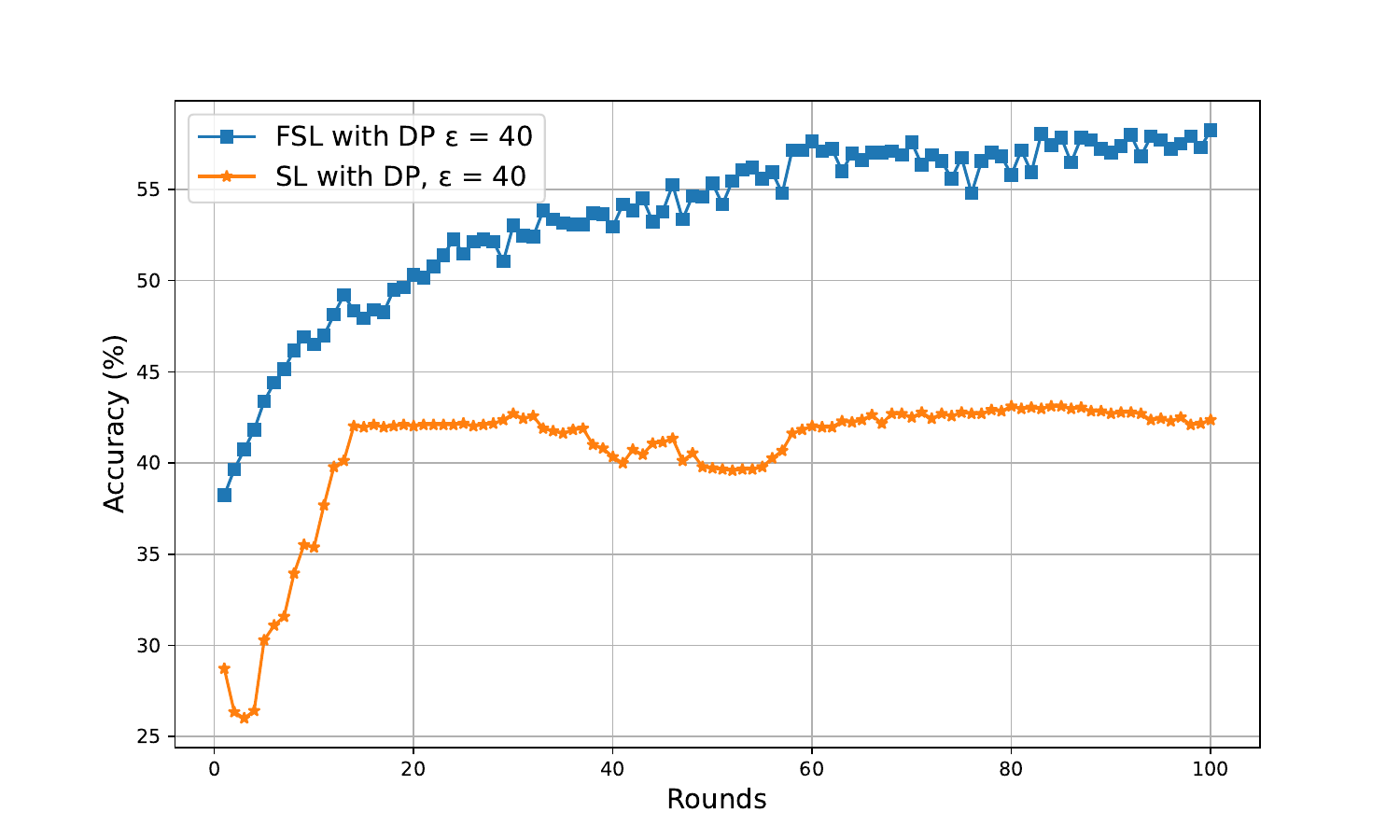}
        \caption{\footnotesize Validation accuracy with  DP.}
        \label{fig4c}
    \end{subfigure}
    \hfill 
    \begin{subfigure}[t]{0.49\linewidth} 
        \centering
        \includegraphics[width=\linewidth]{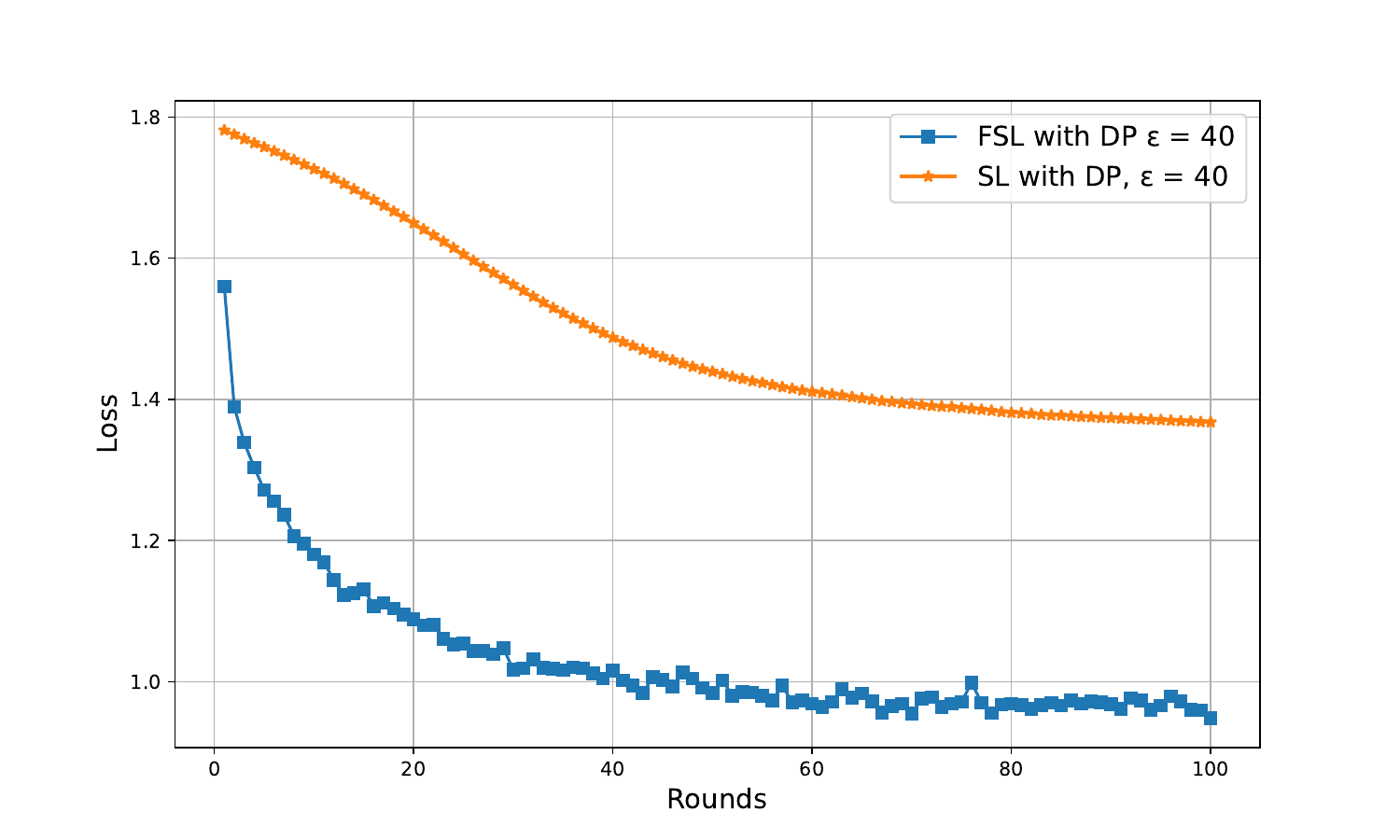}
        \caption{\footnotesize Validation loss with  DP.}
        \label{fig4d}
    \end{subfigure}
    \caption{\footnotesize Comparison between the proposed FSL and traditional FL.}
    \label{fig:three_subs}
\end{figure}

In addition, we examine the performance of traditional FL model as compare to our FSL framework. Figs.~\ref{fig4a} and ~\ref{fig4b} showcase that our framework outperforms traditional FL models. specifically, our FSL framework achieves higher accuracy and lower loss values across 100 rounds of training, demonstrating improved learning efficiency.

In Figs.~\ref{fig4c} and ~\ref{fig4d}, we compare FSL with FL with DP, and $\epsilon$ = 40. We observe that our model once again outperformed traditional FL models with an accuracy increase of 37.51\% and loss decrease of 30.65\%. This indicates that the FSL framework with our design of model splitting and aggregation provides significant advantages over conventional FL, particularly in terms of model convergence and performance stability.

\subsubsection{Comparison of communication time between the proposed FSL and traditional FL}
\begin{figure}[ht!]
    \centering
    \includegraphics[width=0.4\textwidth]{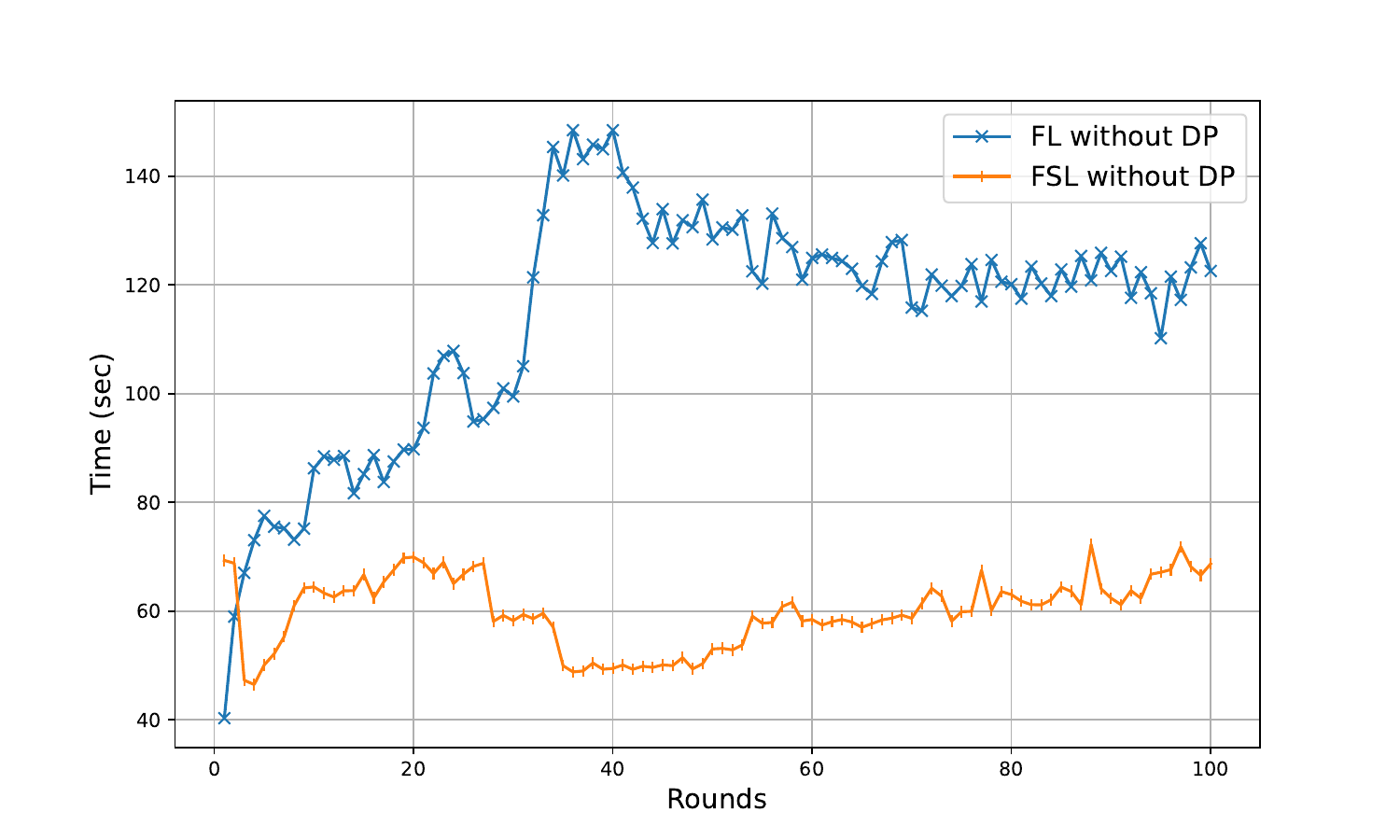} 
    \caption{\footnotesize Comparison of communication time between the proposed FSL and traditional FL.}
    \label{fig5}
\end{figure}

Moreover, we compare the communication time required to train each round between the proposed FSL and traditional FL methods. As shown in Fig.~\ref{fig5}, our FSL framework demonstrates quicker training times per round in more than 95\% of the training rounds. This efficiency in communication time highlights the effectiveness of the FSL framework in reducing latency and improving overall training speed compared to traditional FL. For example, at the global round of 100, our proposed FSL scheme only requires around 65 seconds to complete the entire federated model training, while the FL scheme needs 123 seconds, showing around 100\% training time savings offered by our algorithm design. 

\section{Conclusion} \label{Sec:Conclusion}

This paper has proposed a novel FSL approach for human activity recognition. We have developed a framework using an LSTM model to accurately detect human activity using real-life sensor data. We also implemented a DP mechanism to further enhance privacy protection for activations shared with the server. Simulation results indicate that our proposed FSL framework has outperformed existing FL approaches with better accuracy and loss. Our model also achieved significant latency reductions, with over 100\% lower training time for some rounds compared to traditional FL methods.

\bibliographystyle{ieeetr}
\bibliography{Reference}
\end{document}